\documentclass{article}

\usepackage[english]{babel}

\usepackage{amsmath,amssymb,amsfonts}
\usepackage{graphicx}

\usepackage{geometry}

\usepackage{authblk}
\usepackage[numbers]{natbib}
\usepackage{xcolor}
\usepackage{hyperref}
\hypersetup{
    colorlinks=False,
    linkbordercolor=red,
}
\usepackage{multirow}

\newcommand\correspondingauthor{\thanks{Corresponding author. \\ \textit{e-mail:} ijkim@handong.edu}}

\title{PECI-Net: Bolus segmentation from video fluoroscopic swallowing study images using preprocessing ensemble and cascaded inference}
\author[1,2]{Dougho Park}
\author[3]{Younghun Kim}
\author[3]{Harim Kang}
\author[3]{Junmyeoung Lee}
\author[3]{Jinyoung Choi}
\author[1]{Taeyeon Kim}
\author[1]{Sangeok Lee}
\author[1]{Seokil Son}
\author[1]{Minsol Kim}
\author[3]{Injung Kim\correspondingauthor}    

\affil[1]{Pohang Stroke and Spine Hospital, Pohang, Republic of Korea}
\affil[2]{School of Convergence Science and Technology,
Pohang University of Science and Technology, Pohang, Republic of
Korea }
\affil[3]{School of CSEE, Handong Global University
, Pohang, Republic of Korea}

\date{}

\begin{document}
\maketitle

\begin{abstract}
Bolus segmentation is crucial for the automated detection of swallowing disorders in videofluoroscopic swallowing studies (VFSS).
However, it is difficult for the model to accurately segment a bolus region in a VFSS image because VFSS images are translucent, have low contrast and unclear region boundaries, and lack color information.
To overcome these challenges, we propose PECI-Net, a network architecture for VFSS image analysis that combines two novel techniques: the preprocessing ensemble network (PEN) and the cascaded inference network (CIN).
PEN enhances the sharpness and contrast of the VFSS image by combining multiple preprocessing algorithms in a learnable way.
CIN reduces ambiguity in bolus segmentation by using context from other regions through cascaded inference.
Moreover, CIN prevents undesirable side effects from unreliably segmented regions by referring to the context in an asymmetric way.
In experiments, PECI-Net exhibited higher performance than four recently developed baseline models, outperforming TernausNet, the best among the baseline models, by 4.54\% and the widely used UNet by 10.83\%.
The results of the ablation studies confirm that CIN and PEN are effective in improving bolus segmentation performance.
\end{abstract}

\section{Introduction}

\label{sec:introduction}
Swallowing, an intricate system, is the process of moving food from the lip through the pharynx and into the esophagus \citep{mccarty2021dysphagia}. It involves multiple stages: the oral preparatory, oral transport, pharyngeal, and esophageal phases. The malfunction of one or more of these stages may lead to dysphagia \citep{suttrup2016dysphagia}. Dysphagia is common in brain disorders and is also known to affect around 20-30\% of the older population \citep{takizawa2016systematic, wolf2021prevalence} and can lead to potentially fatal conditions such as malnutrition, dehydration, and aspiration pneumonia \citep{yoon2019correlations}. As a result, dysphagia can significantly deteriorate a patient's long-term prognosis and quality of life \citep{gustafsson1991dysphagia}. Therefore, the accurate evaluation and timely intervention of dysphagia are crucial for the patient's prognosis.

The videofluoroscopic swallowing study (VFSS), the current gold standard for assessing dysphagia, allows real-time evaluation of all swallowing stages from the oral to the esophageal phase \citep{martin2008videofluorographic}. It enables the direct identification of unique issues at each step, as well as the quantification of penetration and aspiration to some extent \citep{tomita2018video}.
However, this gold standard study also comes with a set of challenges. Accurately interpreting VFSS results requires substantial training and experience for human experts; therefore, analyzing the results of VFSS is susceptible to inter-rater variability \citep{silbergleit2018impact}. In addition, doing the VFSS test itself poses a risk of aspiration for the patient \citep{kim2022clinical}. Since both the examiner and the patient are exposed to radiation during the test, deriving the most accurate results with minimal exposure is imperative \citep{pikus2003videofluoroscopic}.

One of the primary objectives of the VFSS examination is to discern the penetration or aspiration during the pharyngeal phase \citep{rosenbek1996penetration}. This phase is characterized by the movement of the food bolus from the oropharynx to the upper esophageal sphincter (UES) via the swallowing reflex. It entails the closure of the nasopharynx and larynx, the hyoid bone and larynx movement in an anterior-superior direction, and the opening of the UES. These coordinated muscular actions occur within a brief span of about one second \citep{zhu2017evaluation}.

Deep learning (DL) has emerged as one of the most vigorously researched areas in the medical field \citep{egger2022medical}. Correspondingly, several studies have been conducted to develop DL models to analyze VFSS; these models are gaining attention as potential alternatives to address the existing limitations of VFSS testing \citep{sejdic2020artificial}. Specifically, it is expected that an automated VFSS interpretation system can enhance the consistency of the results. Additionally, it can verify and reinforce human experts' interpretations and minimize unnecessary exposure to radiation or risk of aspiration. A prerequisite for developing a DL-based system to detect pharyngeal penetration and aspiration is the creation of automated models capable of identifying the bolus and essential structures. However, there has been a lack of research on models that recognize and track the bolus in VFSS images.

\begin{figure}[h]
    \centering
    \includegraphics[scale=.8]{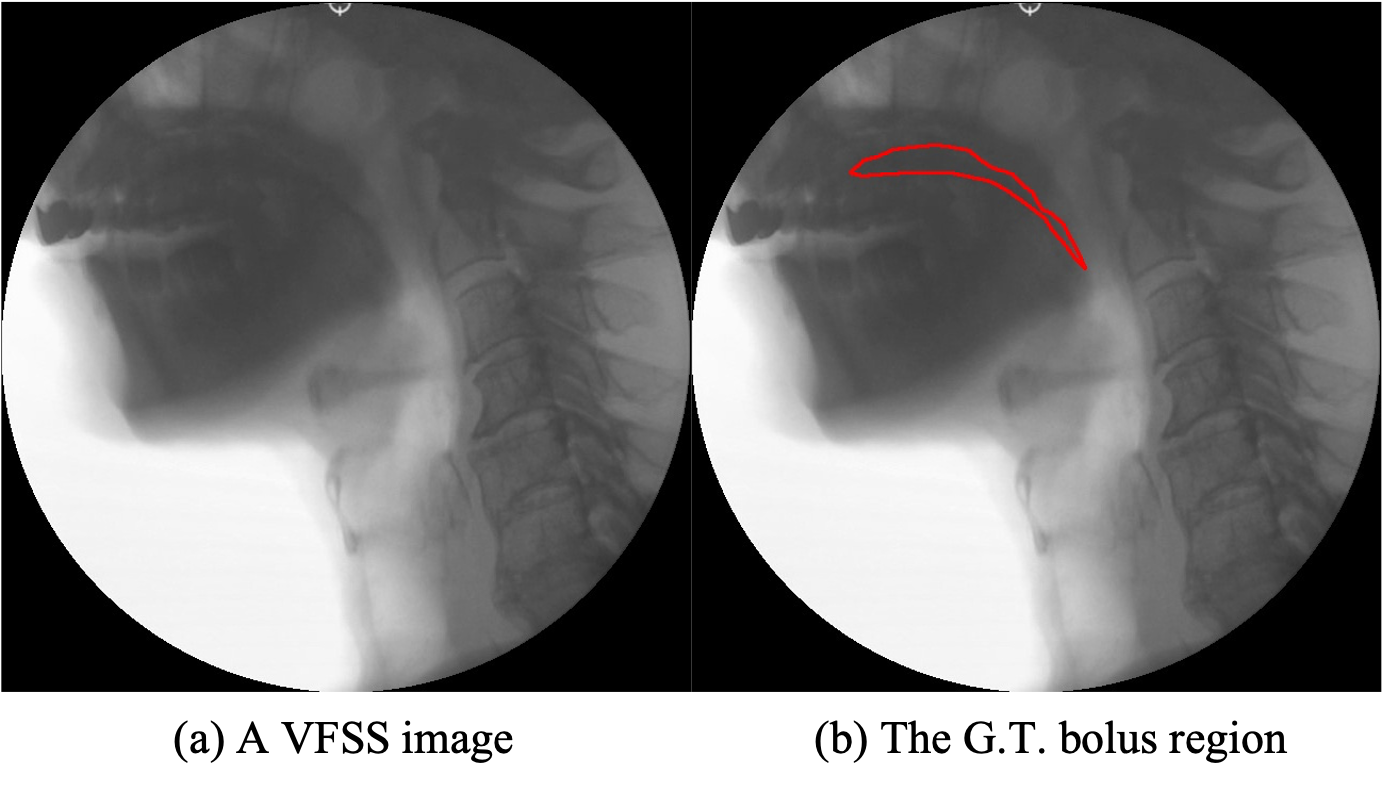}
    \caption{An example of VFSS image with low contrast and the ground truth (G.T.) bolus region.}
    \label{fig:vfss_image}
\end{figure}

The segmentation of bolus from VFSS images is challenging for the following reasons.
First, a bolus may have a liquid or semisolid form that does not have a fixed shape, which introduces significant ambiguity.
Second, unlike ordinary RGB images, VFSS images are translucent, and therefore, each pixel in a VFSS image can belong to more than one organ or object because X-ray sensors detect radiation transmitted through a human body.
As in Fig. \ref{fig:vfss_image}, a pixel in a VFSS image can belong to both a bolus and the surrounding organ, such as the oral cavity, as both are in the trajectory of the radiation arriving at that coordinate.
Third, in VFSS images, many organ areas are dimmed and have unclear boundaries.
This problem is particularly evident in the esophagus or trachea, through which the bolus traverses. The esophagus or trachea provide important clues for bolus segmentation and the diagnosis of swallowing disorder, but due to their high X-ray permeability, they are barely visible in most VFSS images. This issue is compounded by differences in image quality between imaging devices.
Fourth, VFSS images lack color information, requiring the separation of objects and organs based solely on the topography of pixel intensity, which significantly increases ambiguity in bolus segmentation.
In addition, in the VFSS field, privacy, costs associated with imaging and labeling, and other issues hinder the construction of large-scale labeled datasets.
To overcome these difficulties and effectively segment bolus in VFSS images, we need a network architecture and preprocessing algorithm carefully trailed for VFSS image analysis.

Our study aims to overcome the aforementioned challenges and develop a model that can accurately segment bolus in VFSS images, which is a key technology for an automated and accurate penetration and aspiration detection system. To this end, we propose the Preprocessing Ensemble and Cascaded Inference Network (PECI-Net), a novel network architecture specialized for the segmentation of VFSS images. In contrast to the conventional segmentation models that use a softmax output layer to assign each pixel to a single region, PECI-Net estimates the probability of a pixel belonging to each region using a sigmoid output layer. Thus, PECI-Net does not normalize the probability of regions, allowing a pixel to be assigned to multiple regions.

PECI-Net enhances the input image using a preprocessing ensemble network (PEN) to accurately distinguish the regions of bolus and organ in VFSS images with low contrast and unclear region boundaries. PEN constructs a learnable preprocessing module for VFSS image analysis by combining multiple preprocessing algorithms with a neural network block.
To address ambiguity due to the lack of color information and clarity in VFSS images, PECI-Net refers to the global context via a cascaded inference network (CIN). CIN consists of a sequential structure that first segments regions that are reliably detected and strongly related to the location of bolus, and then, referring to the results of the previous stage, segments ambiguous regions, such as bolus. We determine the regions to segment in the first stage based on their importance in bolus segmentation estimated by GradCAM \citep{gradcam} analysis.

In addition, we overcome the scarcity of labeled VFSS image data by transferring the knowledge of a segmentation model pretrained on the ImageNet dataset. To maintain compatibility with the pretrained segmentation model, we match the number of output channels of PEN with the number of input channels of the segmentation model.

To train and evaluate PECI-Net, we built a labeled VFSS image dataset. We recorded 94 VFSS video clips from 47 subjects, cut them into frames, and filtered the frame images according to the criteria list in Subsection \ref{subsec:data_collection_process}. In this way, we collected 2155 VFSS images. Then, rehabilitation physicians and therapists manually labeled the regions of the bolus, mandible, hyoid bone, vocal fold, cervical spine, and soft tissue.

The main contributions of this paper are summarized as 
1) a cascaded inference network (CIN) that improves segmentation accuracy for faintly appearing regions by exploiting the positional correlation between bolus and organs.
2) a preprocessing ensemble network (PEN) to enhance VFSS images by combining multiple image processing algorithms in a learnable way.
3) a labeled VFSS image dataset consisting of 2155 VFSS images and six types of region labels, and a bolus segmentation accuracy of 73.45\%, which is 10.8\% and 4.54\% higher than UNet \citep{unet} and TernausNet \citep{ternausnet}, respectively.

\section{Related Work}
\subsection{Segmentation Models for Medical Applications}
Since the advent of deep learning, convolutional neural networks (CNNs) have been widely used in image processing, including image segmentation. \citeauthor{fcn} performed semantic segmentation using a fully convolutional network. \citeauthor{segnet} proposed SegNet, an image segmentation model with a convolutional encoder-decoder structure. \citeauthor{unet} proposed a network architecture UNet for biomedical image segmentation.  UNet has a symmetric encoder-decoder structure combined with skip connections and has been widely used in subsequent studies on image segmentation.
Transformer \citep{transformer}, which has shown excellent performance in the field of natural language processing since several years ago, has been actively applied in the field of image processing. Dosovitskiy et al. achieved higher recognition performance than CNN on the ImageNet dataset using Vision Transformer (ViT), a transformer-based model for image processing \citep{vit}. \citeauthor{swin} proposed SwinTranformer, a general-purpose backbone network that extends ViT to a multi-scale architecture using shifted windows. 

There have also been studies on applying Transformer to UNet to improve its performance. \citeauthor{swin-unet} proposed SwinUNet, a UNet architecture based on SwinTransformer instead of CNN. \citeauthor{TransUNet} proposed a hybrid encoder, TransUNet, that combines ResNet and Transformer to refer to a broader context. In this work, we implemented each stage of CIN based on TransUNet.

There are a number of studies to build a universal foundation model for image segmentation. \citeauthor{sam} proposed a foundation model for image segmentation called Segment Anything Model (SAM). SAM estimates the segmentation masks of the target object from a prompt. \citeauthor{medsam} adapted SAM to medical image segmentation using more than 1M medical image-mask pairs. Their model, MedSAM, outperformed SAM and UNet on several medical image segmentation benchmarks. Although SAM and MedSAM demonstrated the potential of foundation models for image segmentation, they require manually specified prompts, such as points and bounding boxes, and are not applicable for fully automated segmentation.

\subsection{Bolus Segmentation and Detection}
There is a notable scarcity of related literature concerning the segmentation and detection of a food bolus from VFSS images.  \citeauthor{caliskan2020automated} attempted automatic bolus detection in VFSS using a Mask R-CNN \citep{maskrcnn} model. Their model showed a performance of 0.71 in intersection over union (IoU), demonstrating the possibility of automatic bolus tracking from VFSS images despite several limitations. \citeauthor{ariji2022preliminary} utilized the widely used UNet model for medical image segmentation and demonstrated a performance of Dice score 0.94±0.05.

\citeauthor{zeng2023video} introduced a VFSS video analysis model, VideoTransUNet, which integrates the Temporal Context Module (TCM) to consider the temporal characteristics of video frames within the TransUNet \citep{TransUNet} architecture. This incorporation led to a notable improvement, with the model achieving a Dice score of 0.8796. \citeauthor{VideoSwinUNet} devised VideoSwinUNet, which blends the temporal features of video frames using the more powerful SwinTransformer instead of ViT, increasing the Dice score to 0.8986. However, their dataset was not fully released to the public for ethical reasons.

\citeauthor{iida2023detection} directly detected aspiration from VFSS images using CNN-based classification models. They set a region of interest on the pharynx and detected aspiration using a simple-layer CNN, multi-layer CNN, and modified LeNet.
Since none of the existing research makes their data publicly available, it is difficult to directly compare their performance.

\subsection{Hyoid Bone Segmentation and Detection}
Several studies have proposed automated models for detecting and tracking the hyoid bone, a critical element in swallowing studies \citep{kim2017semi, feng2021automatic, hsiao2023deep}. \citeauthor{zhang2018automatic} presented a model that automatically detects the hyoid bone in a frame. Their single-shot multi-box detector model was able to locate the hyoid bone with an average precision of 89.14\%. In addition, they successfully localized cervical vertebral bodies using a two-stage convolutional neural network model in their other study \citep{zhang2021automatic}. \citeauthor{kim2021hyoid} proposed a system that combined a bidirectional feature pyramid network and a bottleneck Transformer \citep{transformer} to detect hyoid bone movement. Their model demonstrated an area under the receiver operating characteristics curve (AUROC) of 0.998 for pixel-wise accuracy. The hyoid bone, linked to the suprahyoid muscle, plays a critical role in airway closure during the pharyngeal phase and the opening of the UES. As such, the movement of the hyoid bone is intricately associated with these processes. The significance of these studies lies in their demonstration of the potential applicability of an automatic hyoid bone tracking system in VFSS interpretation. These findings represent valuable advancements in automating and enhancing the accuracy of dysphagia assessment.

\subsection{Penetration Classification and Swallowing Delay Measure}
\citeauthor{kim2022deep} presented a model that detected normal swallowing, penetration, and aspiration by applying a convolutional neural network to VFSS images obtained from 190 patients and showed a performance of 0.940 of macro average AUROC.
Meanwhile, there have been studies to evaluate abnormalities in the swallowing process. Two studies reported an attempt to detect the pharyngeal swallowing reflex time by presenting a DL model that recognizes a specific stage (pharyngeal phase) in VFSS images, which can enable quantitative evaluation of findings such as swallowing reflex delay \citep{lee2021automatic, lee2020machine}.

\subsection{Sequential Architectures}
\citeauthor{ramakrishna2014pose} proposed Pose Machines (PM) to address the limitations of the existing human pose estimators that fail to capture interactions between body parts. \citeauthor{wei2016convolutional} proposed Convolutional Pose Machines (CPM) to implicitly model long-range dependencies between key points in a prediction task by multi-stage Pose Machines. CPM devises an effective receptive field to enable the model to learn long-range spatial dependencies and improve performance through a side-by-side convolutional architecture. In addition, they reduce the gradient vanishing problem during training by adding an intermediate loss layer after each stage.
In object detection, \citeauthor{cascade-R-CNN} proposed a multi-stage object detector, Cascade R-CNN, to overcome the problem of training with a single IoU threshold. They improved precision without sacrificing recall by concatenating multiple detectors trained with increasing IoU thresholds.

\begin{figure*}[t]
    \centering
    \includegraphics[width=\textwidth]{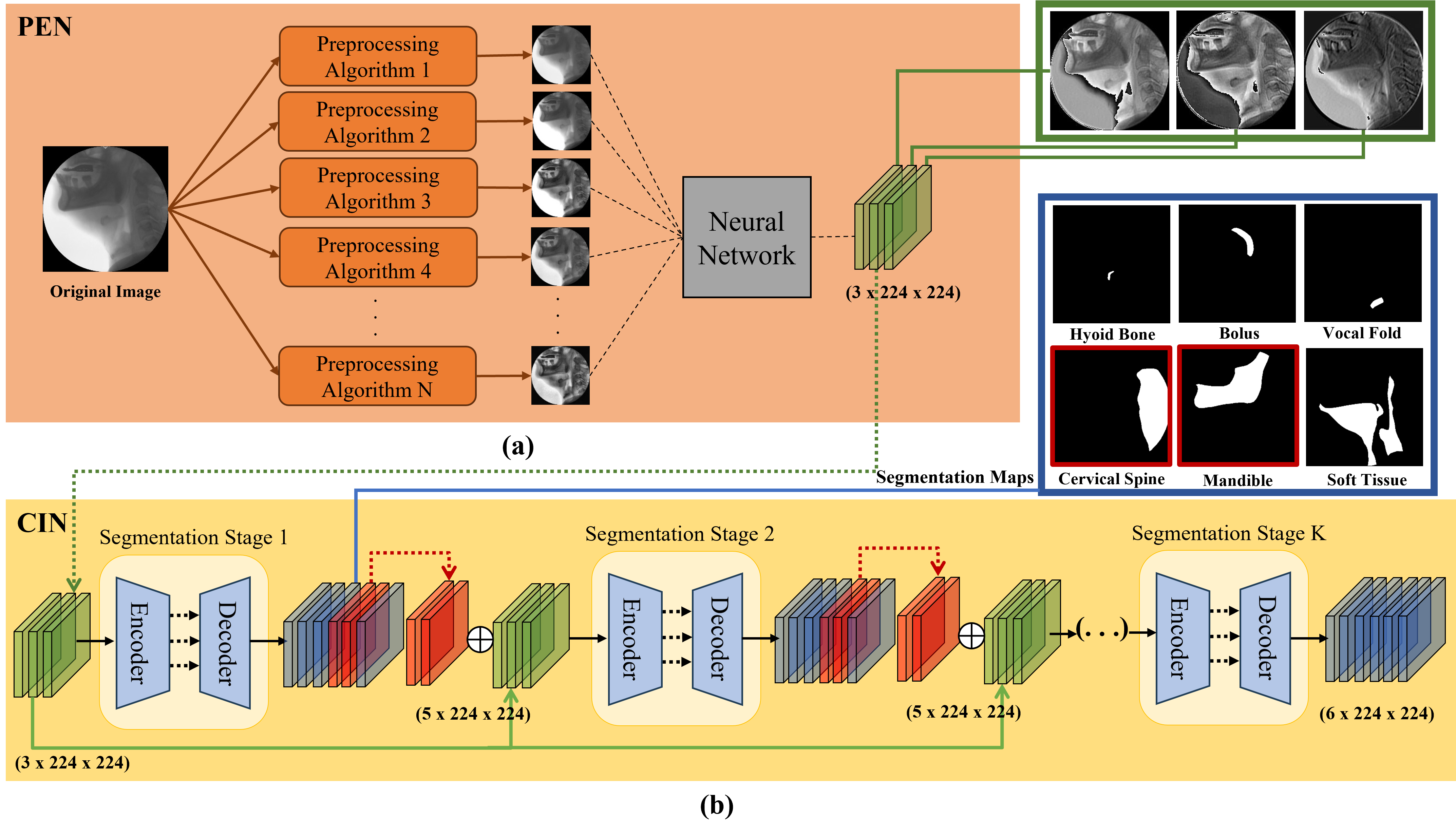}
    \caption{The architecture of PECI-Net. (a) Preprocessing Ensemble Network (PEN); (b) Cascaded Inference Network (CIN) }
    \label{fig:overall-architecture}
\end{figure*}

\section{Preprocessing Ensemble and Cascaded Inference Network}

\subsection{Overall Architecture}
As shown in Fig. \ref{fig:overall-architecture}, the proposed PECI-Net consists of a Preprocessing Ensemble Network (PEN) that enhances VFSS images by combining multiple preprocessing algorithms and a Cascaded Inference Network (CIN) that segments VFSS images by utilizing the positional relationship between regions through a sequential structure. PEN applies multiple image processing algorithms to the input image and then combines their results into three channels using a neural network block as Fig. \ref{fig:overall-architecture}(a). In training, the parameters of the neural network block are optimized to minimize a segmentation loss. Therefore, PEN learns the optimal combination of image processing algorithms to maximize bolus segmentation performance. The number of output channels of PEN is the same as the number of channels of ordinary RGB images, which maintains compatibility with pretrained segmentation models and thereby makes it easier to take advantage of transfer learning.

CIN takes the three-channel enhanced image as input and segments the regions of the bolus and organs in stages. CIN comprises a series of segmentation modules, as shown in Fig. \ref{fig:overall-architecture}(b).
In the first stage, it segments the input image, focusing on the regions that are detected reliably and have a close spatial relationship with the bolus. Based on the segmentation results for those regions, the subsequent stages segment ambiguous regions, such as bolus, more accurately. Since each step generates segmentation maps using a sigmoid layer, each pixel can be assigned to more than one region. 

In this study, we implemented the segmentation stages based on TransUNet \citep{TransUNet}. As it is hard to collect a sufficient amount of labeled VFSS images, we improved training efficiency by transferring the parameters pretrained on the ImageNet dataset.

\begin{figure}[h]
    \centering
    \includegraphics[width=0.7\textwidth]{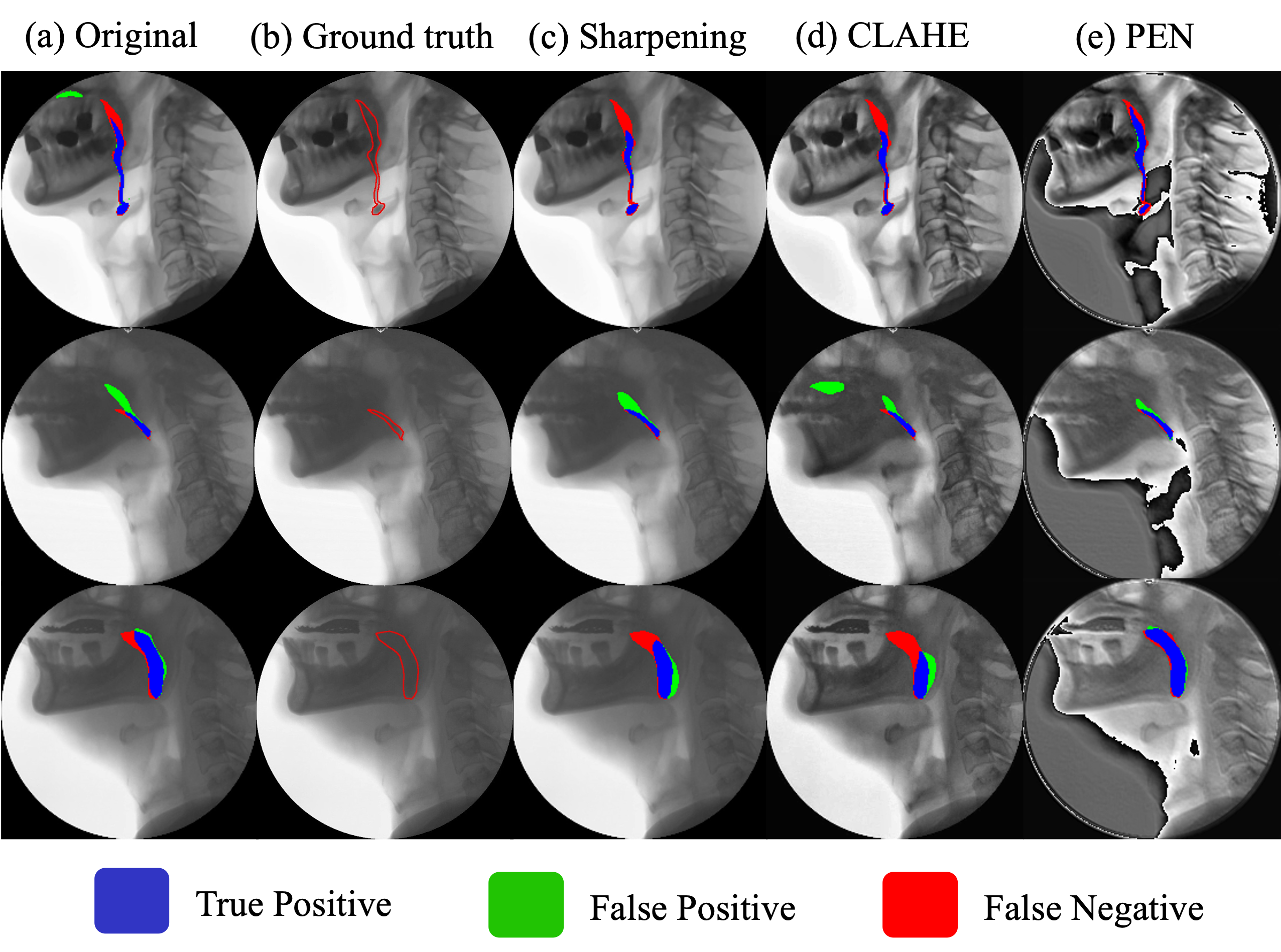}
    \label{fig:preprocess_comparison_figure}
    \caption{ The enhanced images and bolus segmentation results according to the preprocessing algorithm. (a) input image, (b) G.T. bolus region, (c) Laplacian sharpening, (d) CLAHE, and (e) PEN. (PEN outputs a three-channel enhanced image as in Fig. \ref{fig:preprocessing_ensemble}. (e) displays the average of them to save space.)}
    \label{fig:comparison_of_preprocessing}
\end{figure}

\subsection{Preprocessing Ensemble Network}
\label{subsec:PEN}
X-ray images are widely used in medical diagnosis, including VFSS. However, X-ray commonly produces medium-quality images with high noise, low intensity, poor contrast, and weak region boundaries \citep{mustafa2020overview}. 
The limited image quality obstructs the precise segmentation of the bolus and organs.
Fig. \ref{fig:comparison_of_preprocessing}(a) displays a few examples of VFSS images and the bolus region estimated by CIN. The blue color indicates correctly classified pixels, while the green and red colors indicate false positive and false negative pixels, respectively. In these images, the boluses are placed in the mandible regions, and their differences in pixel intensity values are insignificant. This results in significant segmentation errors at the unclear boundaries. 

When the bolus is located across a dark and light background, as in the top of Fig. \ref{fig:comparison_of_preprocessing}, accurate bolus segmentation is even more difficult. To overcome this challenge, a powerful preprocessing algorithm is essential.
Previous studies have proposed various image processing techniques to enhance the contrast and sharpness of digital images. Among them, Laplacian sharpening \citep{gonzales18} is widely used to enhance the detail of images and make region boundaries sharper. Another line of preprocessing algorithms is histogram equalization (HE) \citep{histogram_equalization}, which redistributes the intensity histogram of the input image to increase the contrast between regions and improve visibility. However, when applied to images with uneven intensities, equalizing the global histogram causes undesirable side effects, such as increasing background noise or overly emphasizing uninterested parts. Adaptive histogram equalization (AHE) mitigates this issue by dividing the input image into contextual regions and redistributing the pixel intensities based on local intensity histograms. They apply bilinear interpolation to avoid discontinuity at the boundary between contextual regions. However, AHE often produces noisy results in homogeneous areas. Contrast limited adaptive histogram equalization (CLAHE) \citep{CLAHE} alleviates the noise problem by limiting contrast enhancement.

While these algorithms enhance image quality in some aspects, applying one of them alone is sub-optimal for the segmentation of VFSS images for the following reasons. Firstly, each image processing algorithm has its own strengths and weaknesses, and it is hard to develop a single algorithm that is generally best for all images. Second, although the preprocessing algorithms generate images that look better to the human eye, they are handcrafted algorithms and lack training methods to optimize them for a specific task, such as bolus segmentation. Third, whereas most segmentation models take a three-channel image as input, the preprocessing algorithms only produce a single-channel image when applied to a gray-scale image, thus utilizing only a fraction of the bandwidth. Consequently, both Laplacian sharpening and CLAHE provide limited improvements when applied alone, as shown in Fig. \ref{fig:comparison_of_preprocessing}(c) and (d), where significant segmentation errors remain.

PEN overcomes these limitations by combining multiple image processing algorithms with a neural network block and optimizing the network parameters for bolus segmentation, as illustrated in Fig. \ref{fig:overall-architecture}(a). Thus, PEN learns the optimal combination of multiple preprocessing algorithms for VFSS image segmentation. First, it generates enhanced images by applying multiple image processing algorithms as Eq. \ref{eqn:preprocessing}.
\begin{equation}
    \label{eqn:preprocessing}
    x_i = Preprocess_i(x), \text{ for } 1 \leq i \leq N,    
\end{equation}
where $N$ is the number of image processing algorithms.

Then, PEN concatenates the enhanced images to form a $N \times\ H \times W$ feature map, where $H \times W$ is the size of the input image. Finally, it compresses the concatenated feature maps into three channels using a neural network block, producing a $3 \times H \times W$ feature map. In this study, we compress the enhanced images using a $7 \times 7$ convolution followed by a ReLU activation as Eq. \ref{eqn:PEN}.
\begin{equation}
    \label{eqn:PEN}
    \overline{x} = PEN(x) = ReLU(Conv_{7\times7}(Concat(x_1, \ldots, x_N))),
\end{equation}
where the output feature map $\overline{x} \in \mathbb{R}^{3 \times H \times W}$ is fed to the segmentation model.

In this study, we combine the following five image-processing algorithms:
\begin{itemize}
    \item Identity mapping ($x_1 = x$)
    \item Laplacian sharpening ($x_2 = Sharpen(x)$) 
    \item CLAHE ($x_3 = CLAHE(x)$)
    \item Double CLAHE ($x_4 = CLAHE(CLAHE(x))$)
    \item CLAHE + Sharpening ($x_5 = Sharpen(CLAHE(x))$)
\end{itemize}

Since image processing algorithms can partially damage meaningful information from the original image, we added the identity mapping to ensure that all information in the input image is delivered to the neural network block. Laplacian sharpening enhances the clarity of region boundaries by adding Laplacian to the intensity of each pixel as Eq. \ref{eqn:sharpening} \citep{gonzales18}.

\begin{equation}
\begin{split}
    y(i,j) = & \text{ } 5 x(i,j) - [x(i+1,j) + x(i-1, j)  \\ 
             & + x(i,j+1) + x(i,j-1)].  
             \label{eqn:sharpening}
\end{split}
\end{equation}

CLAHE divides the input image into subregions, computes the intensity histogram in each subregion, clips the histograms using a predefined threshold, and transforms the intensity of each pixel to equalize local intensity histograms \citep{CLAHE}. To avoid discontinuity at subregion boundaries, it transforms each pixel using the intensity histograms of the nearest four subregions via bilinear interpolation.
CLAHE enhances the contrast of the input image, which helps separate the bolus from the surrounding organs, such as the mandible and soft tissues. We also added two composite algorithms: Double CLAHE, which applies CLAHE twice to enhance contrast further, and the combination of CLAHE and Laplacian sharpening to enhance both clarity and contrast.

In training, we combine PEN to the segmentation model, CIN, and jointly optimize their parameters, $\theta_{PEN}$ and $\theta_{CIN}$, to minimize the segmentation loss defined as Eq. \ref{eqn:total_loss}. The gradients are computed by backpropagation as Eq. \ref{eqn:gradient_of_PEN}, where $\frac{\partial TotalLoss}{\partial \overline{x}}$ is delivered from the segmentation model.
\begin{equation}
    \frac{\partial TotalLoss}{\partial \theta_{PEN}} = \frac{\partial \overline{x}}{\partial \theta_{PEN}} \frac{\partial TotalLoss}{\partial \overline{x}}
    \label{eqn:gradient_of_PEN} 
\end{equation}

\begin{figure}[h]
    \centering    
    \includegraphics[width=0.5\textwidth]{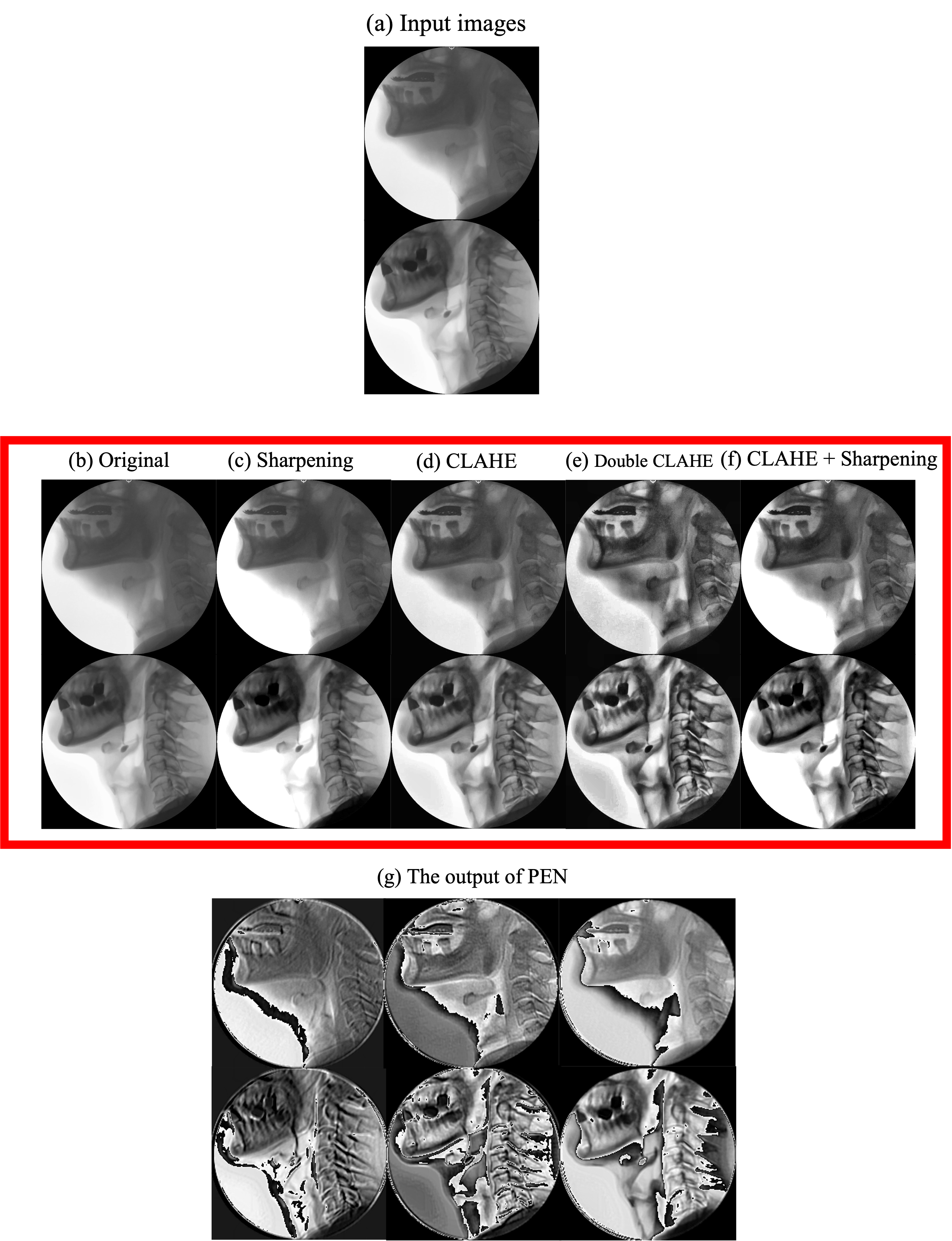}  
    \caption{Preprocessing ensemble. (a) input images, (b)-(f): the results of preprocessing algorithms, (g) the output of PEN (3 channels).}
    \label{fig:preprocessing_ensemble}
\end{figure}

Fig. \ref{fig:preprocessing_ensemble} shows the process of PEN by displaying the input images, the results of the five preprocessing algorithms, and the three-channel outputs of PEN.
The output channels show improved clarity in region boundaries and increased contrast. Moreover, the three channels exhibit different characteristics, suggesting that each channel learns an image transform complementary to each other.

\subsection{Cascaded Inference Network for Medical Image Analysis}

\subsubsection{Cascaded Inference}

As described in Section \ref{sec:introduction}, the major challenges in bolus segmentation in the VFSS image include ambiguity due to the non-fixed shape of the bolus, the translucency and insufficient quality of VFSS images, and the absence of color information. For example, the bolus region in Fig. \ref{fig:vfss_image} has a similar intensity to the surrounding oral cavity, which makes it difficult to segment correctly.
We reduce the ambiguity by using the spatial relation between the bolus and organs. During swallowing, the bolus traverses along a path that leads to the oral cavity, pharynx, and esophagus. In cases of penetration and aspiration, the bolus may enter the trachea. This provides a strong prior for bolus segmentation.
However, it is not straightforward to utilize such prior for narrowing down the candidate bolus location because the spatial relationships between the bolus and organs are complex, and there are also ambiguities in detecting the organs around the bolus' path. 

In the field of human pose estimation, Pose Machine (PM) \citep{ramakrishna2014pose} and Convolutional Pose Machine (CPM) \citep{wei2016convolutional} reduce ambiguities by referring to the context through cascaded inference. They implicitly learn potentially complex spatial relations between human body parts using a sequential architecture. Inspired by these studies, we reduce ambiguity in bolus segmentation referring to the context by cascaded inference.
CIN is composed of a series of segmentation stages, as shown in Fig. \ref{fig:overall-architecture}(b). Each stage, except for the first one, takes as input the segmentation result of the previous stage, in addition to the input image enhanced by PEN. As a result, they can segment the bolus and organs more accurately by referring to the context from the preliminary segmentation results of other regions.

Besides modifications to fit bolus segmentation, CIN extends the concept of cascaded inference in a few ways.
While each stage of PM was implemented with boosted random forests and handcrafted feature extractors and that of CPM is implemented with convolution operators, CIN references the global context through Transformer blocks \citep{transformer}.
We constitute CIN based on TransUNet \citep{TransUNet}, which extends the widely-used UNet \citep{unet} by combining Transformer blocks at the end of the encoder. As a result, CIN can reference the global context at each stage, which is important for learning spatial relations between bolus and distant organs, such as the cervical spine.

More importantly, CIN avoids undesirable side effects from the wrong context by referencing contexts in an asymmetric way.
Unlike the portrait image, in the VFSS image, each region has a different opacity and clarity. While some regions are segmented reliably, others are difficult to segment. The context from incorrectly segmented regions can harm the prediction of subsequent stages. Therefore, each stage of CIN only refers to the segmentation results for the regions that are reliably detected and closely related to the location of the bolus.

\begin{figure}[h]
    \centering
    \includegraphics[width=0.7\columnwidth]{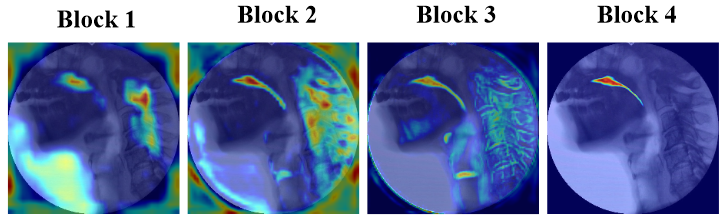}
    \caption{ GradCAM of the four decoder blocks of TransUNet. Red represents the highest values, and purple represents the lowest values. The pixels in the cervical spine and mandible regions are assigned with high importance values. }
    \label{fig:gradcam_TransUNet}
\end{figure}

To select the regions to segment in the first stage, we analyzed the importance of each region in bolus segmentation using GradCAM \citep{gradcam}. We trained a TransUNet model for bolus segmentation and visualized the importance map of the four decoder blocks as Fig. \ref{fig:gradcam_TransUNet}. In general, the pixels in the cervical spine and mandible regions are assigned with high importance values as in Fig. \ref{fig:gradcam_TransUNet}, suggesting that the intermediate feature maps of the two regions provide useful context for bolus segmentation. Additionally, the cervical spine and mandible regions have high opacity and therefore are more reliably segmented than other regions in most VFSS images.
Therefore, we chose the cervical spine and mandible as the context for bolus segmentation in the second stage.
However, the accuracy of the first stage decreased when we trained it only to segment the cervical spine and mandible regions, ignoring other regions. We believe the reason is that removing the other regions from the segmentation targets results in a loss of contextual information for correctly segmenting the cervical spine and mandible regions. Therefore, we train the first stage to segment all of the six regions but transmit only the segmentation maps of the cervical spine and mandible to the next stage. The next stage re-segments the six regions in the enhanced input image based on the segmentation maps received from the first stage. While CIN can consist of any number of stages, in our experiment, the 2-stage model showed the best performance, as described in Subsection \ref{subsec:no_stages} and Table \ref{tab:performance_by_stages}

\subsubsection{CIN Architecture}

As shown in Fig. \ref{fig:overall-architecture}(b), CIN comprises a series of segmentation stages. The first stage $f_1(\cdot)$ segments the input image as Eq. \ref{eqn:stage1}.
\begin{align}
    \hat{a}_1 &= f_1(\overline{x}) = f_1(PEN(x))    \\
    \hat{y}_1 &= \sigma(\hat{a}_1),
    \label{eqn:stage1}
\end{align}

where $\overline{x}$ is the enhanced image generated by PEN as Eq. \ref{eqn:PEN}, and $\hat{y}_1 \in \mathbb{R}^{|T| \times H \times W}$ is the segmentation result of the first stage. $T$ is the set of target regions and $\sigma(\cdot)$ denotes sigmoid activation. In this study, $T$ = \{ bolus, mandible, hyoid bone, vocal fold, cervical spine, soft tissue \}.

The subsequent stages $f_i(\cdot)$, for $i \ge 2$, concatenate $\overline{x}$ and the segmentation result of the previous stage and re-segment all regions from the combined feature map. They use only the selected channels $T' \subseteq T$ and ignore the others to avoid side effects from unreliably segmented regions.
\begin{align}
    \hat{a}_i &= f_i(Concat(\overline{x}, \hat{a}_{i-1}[T'_{i-1}])), \text{ for } T' \subseteq T    \\
    \hat{y}_i &= \sigma(\hat{a}_i),
    \label{eqn:stage_i}
\end{align}
where $\hat{a}_{i-1}[T'_{i-1}]$ denotes the channels of $\hat{a}_{i-1}$ for the selected regions $T'$. In this study, $T'_1$ = \{cervical spine, mandible \}.

In the final stage, we convert the logit to probability estimation by sigmoid activation as in Eq. \ref{eqn:stagen}.
\begin{equation}
    CIN(\overline{x}) = CIN(PEN(x)) = \hat{y}_S = \sigma(\hat{a}_S),
    \label{eqn:stagen} 
\end{equation}
where $S$ is the number of stages. In contrast to most segmentation models that apply softmax activation, CIN does not normalize the probability for the target regions, allowing each pixel to belong to multiple regions.

Each stage $f_i(\cdot)$ consists of a TransUNet \citep{TransUNet}. TransUNet extends the encoder-decoder architecture of UNet \citep{unet}, which is widely used in medical image analysis, by combining 12 Transformer blocks at the end of its encoder to refer to the global context. The input resolution of the first stage is $3 \times 224 \times 224$. The second stage takes as input a $5 \times 224 \times 224 $ feature map, including the segmentation maps of the cervical spine and mandible.
The architecture and hyperparameters of each stage are presented in Table \ref{tab:architecture}.
\begin{table}[h]
\centering
\caption{The architecture and hyperparameters of each stage.}
\label{tab:architecture}

\begin{tabular}{lll}
\hline
Operators             & Kernel Size/Stride & Feature Map Size \\ \hline
ResNet-50             &                    & 14x14x768        \\
Transformer block x12 &                    &                  \\
Layer Norm            &                    & 196x768          \\ \hline
Convolution           & 3x3/1              & 14x14x512        \\
Batch Norm + ReLU     &                    &                  \\ \hline
Upsampling (bilinear) &                    & 28x28x512        \\ \hline
Convolution           & 3x3/1              & 28x28x256        \\
Batch Norm + ReLU     &                    &                  \\ \hline
Upsampling (bilinear) &                    & 56x56x256        \\ \hline
Convolution           & 3x3/1              & 56x56x128        \\
Batch Norm + ReLU     &                    &                  \\ \hline
Upsampling (bilinear) &                    & 112x112x128      \\ \hline
Convolution           & 3x3/1              & 112x112x64       \\
Batch Norm + ReLU     &                    &                  \\ \hline
Upsampling (bilinear) &                    & 224x224x64       \\ \hline
Convolution           & 3x3/1              & 224x224x16       \\
Batch Norm + ReLU     &                    &                  \\ \hline
Convolution           & 3x3/1              & 224x224x6       
\end{tabular}%
\end{table}

\subsection{Training of CIN}
\label{subsec:training}

We train CIN using the Dice loss as Eq. \ref{eqn:total_loss}-\ref{eqn:dice_loss}, which is known to be effective in segmenting small regions \citep{zou2004statistical}.
\begin{align}
    TotalLoss &=  \sum_{i=1}^{S}{StageLoss_i(\hat{y}_i[t], y[t])} \label{eqn:total_loss} \\
    StageLoss_i &=  \sum_{t=1}^{|T|} {w_{i,t} DiceLoss_{i, t} (\hat{y}_i[t], y[t])} \label{eqn:stage_loss}  \\
    DiceLoss_{i,t} &= 1 - {2*\sum_{j=1}^{H \times W}{\hat{y}_i[t][j] \cdot y[t][j]} \over \sum_{j=1}^{H \times W} {\hat{y}_i[t][j]} + \sum_{j=1}^{H \times W} {y[t][j]}} \label{eqn:dice_loss},
\end{align}
where $\hat{y}_i$ denotes the segmentation result of the $i$-th stage and $y$ is the ground truth segmentation maps. $\hat{y_i}[t][j]$ denotes the $j$-th element in the $t$-th channel of $\hat{y_i}$. $w_{i,t}$ is the weight for the $t$-th channel in the $i$-th stage. $DiceLoss_{i,t}$ denotes the per-channel Dice loss, which represents the segmentation accuracy of a region. $StageLoss_i$ accumulates the Dice loss of all channels as a weighted sum, and $TotalLoss$ accumulates $StageLoss_i$ for all stages. The stage loss $StageLoss_i$ for $i < S$ was introduced for intermediate supervision to mitigate the gradient vanishing problem.

In the first stage, we set $w_{1,t} = 1$ for all regions to learn useful contexts without bias. However, in the second stage, we focus on bolus by assigning a higher weight than other regions. In this study, we set $w_{2,t}$ to 2.5 for the bolus and 0.7 for the other regions.

For the first stage, we transfer the parameters of a TransUNet model pretrained on the ImageNet dataset to compensate for the scarcity of labeled VFSS images and improve performance. However, we train the second stage from scratch because the number of input channels differs from that of the vanilla TransUNet.

\section{Data Collection}

\begin{table}[h]
\centering
\caption{Baseline characteristics of patients}
\label{tab:characteristic_of_patients}
\resizebox{0.6\textwidth}{!}{%
\begin{tabular}{lll}
\hline
                                          & Variables              & Values      \\ \hline
Age                                       &                        & 65.6 ± 13.6 \\ \hline
\multirow{2}{*}{Gender (n=47)}            & Male                   & 45          \\
                                          & Female                 & 2           \\ \hline
\multirow{5}{*}{Primary diagnosis (n=55)} & Ischemic stroke        & 42 (76.4)   \\
                                          & Hemorrhagic stroke     & 7 (12.7)    \\
                                          & Traumatic brain injury & 1 (1.8)     \\
                                          & High cervical leison   & 2 (3.6)     \\
                                          & Others                 & 3 (5.5)     \\ \hline
\multirow{3}{*}{Food forms (n=94)}        & Solid                  & 26 (27.7)   \\
                                          & Semisolid              & 27 (28.7)   \\
                                          & Liquid                 & 41 (43.6)   \\ \hline
\end{tabular}%
}
\end{table}

\subsection{Data Collection Process}
\label{subsec:data_collection_process}
\begin{figure}[h]
    \centering
    \includegraphics[width=0.7\textwidth]{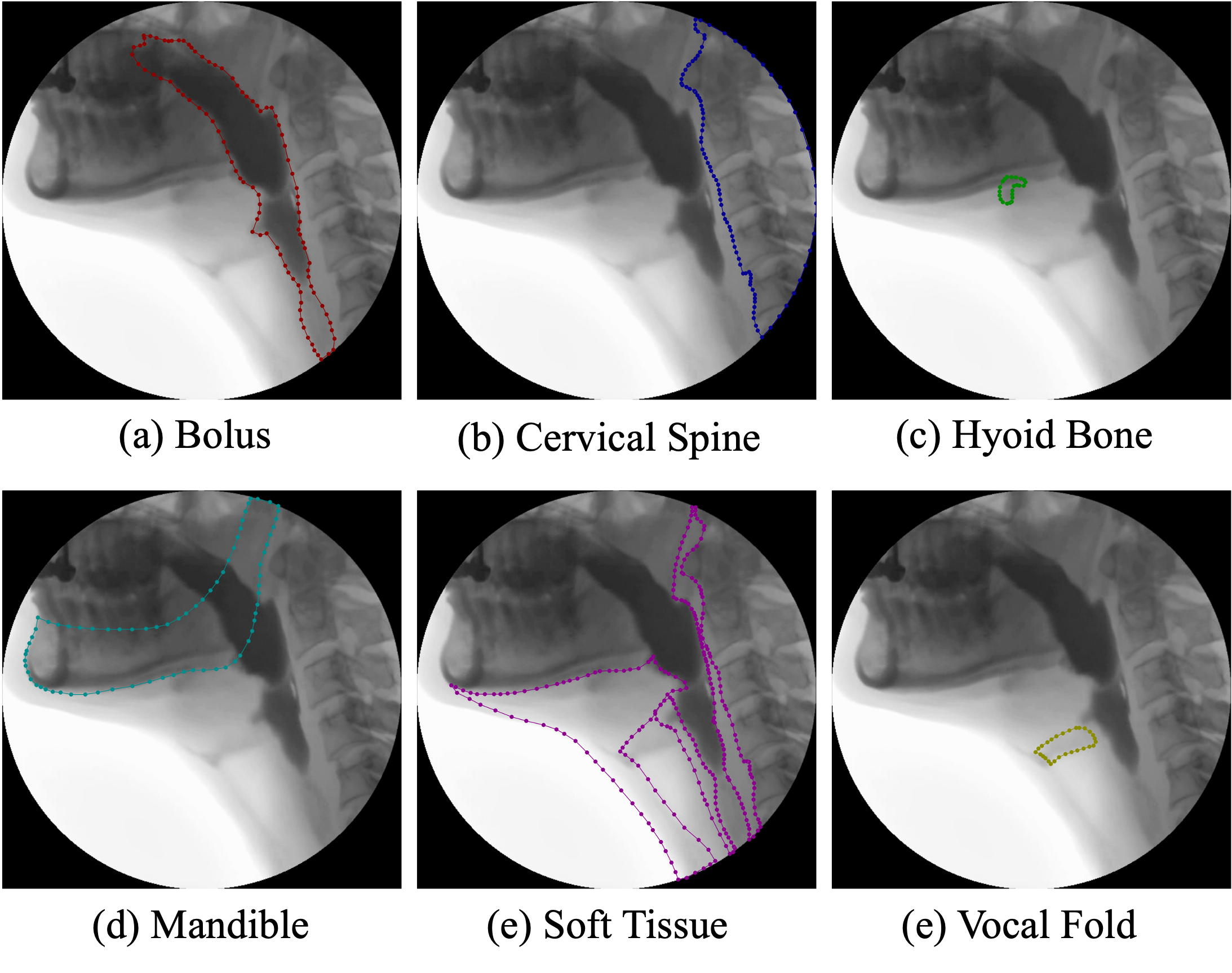}
    \label{fig:annotations}
    \caption{Examples of the ground truth labels of the six target regions}
    \label{fig:target_regions}
\end{figure}
We collected VFSS image data from a single general hospital from December 2021 to April 2023. Finally, we extracted 94 swallowing examinations from a total of 47 patients. The characteristics of the patients are presented in Table \ref{tab:characteristic_of_patients}.
We excluded the following cases: 1) poor image quality, 2) no identified pharyngeal phase during VFSS study, 3) previous head and neck cancer, and 4) lack of clinical information.
The poor image quality indicates images with improper patient posture during VFSS examination or containing motion blurs due to the movement of patients or imaging equipment. We also excluded images with severe blur due to insufficient temporal resolution of the recording device. Furthermore, images challenging to recognize the boundaries of the bolus with the human eye due to insufficient contrast agents were also excluded. Experienced physicians (D.P. and S.L.) identified and excluded the images as poor quality among eligible VFSS examinations. The entire video clips and frames of the final 47 subjects were 95 and 20,562, respectively.
Then, we extracted the pharyngeal phase from VFSS videos and finally utilized 2155 frame images for analysis. Finally, we manually annotated the regions of the bolus, hyoid bone, vocal fold, cervical spine, mandible, and other soft tissue in each image, as shown in Fig. \ref{fig:target_regions}.
The experienced multidisciplinary rehabilitation team consisting of rehabilitation medicine specialists, a speech-language therapist, and occupational therapists participated in the manual annotation. Each annotated image was cross-checked by the team members who co-authored this study. They all have at least five or more years of experience conducting and interpreting the VFSS examination.


\subsection{VFSS Protocol}
The VFSS examination was conducted with the patient in a sitting upright position. The two C-arm fluoroscopy equipment units were used throughout the research period: the ZEN-5000 and the OSCAR Classic (Genoray Inc., Seongnam, Korea). The contrast agent was a barium solution diluted to 35\% in the diet from its original concentration of 230\%. A variety of foods were used for the study, including liquid (2ml, 5ml, and 90ml of water), a spoon of semi-solid (plain yogurt and white rice porridge), and a spoon of solid foods (white rice). The evaluation process was conducted collaboratively by a team consisting of a rehabilitation medicine specialist, a speech-language pathologist, an occupational therapist, and a radiologic technologist. The videos were recorded at 30 frames per second in a sagittal view with a 1280 x 720 resolution.

\subsection{Ethical Statements}
This study protocol was approved by the Institutional Review Board of Pohang Stroke and Spine Hospital (approval number: PSSH0475-202208-HR-012-01). Informed consent for this study was waived due to the retrospective nature of the study design and anonymized image data. Only authorized researchers (D.P. and T.K.) were allowed to access original video data to extract and anonymize each frame. We followed the declaration of Helsinki during the entire study period.

\section{Experiments}

\subsection{Experimental Settings and Evaluation Metric}

We divided the VFSS image data into training, validation, and test sets at a ratio of 8:1:1. We split the datasets based on patient ID to prevent overlap between data sets.
We trained the model for 250 epochs using the AdamW optimizer \citep{adamW} with a batch size of 16. Initially, we set the learning rate to 0.001 and decreased it linearly at each epoch.
We conducted experiments on a server with Zeon E5-2630v4 2.2GHz CPU, 64GB RAM, and an NVIDIA GTX-1080 GPU with 8 GB memory.

For the evaluation metric, we used the Dice score, which is also known as Dice similarity coefficient (DSC) and is widely used for evaluating the accuracy of segmentation models for medical images \citep{dicescore}. We first converted the estimated segmentation maps $\hat{y} \in [0,1]^{|T| \times H \times W}$ to binary prediction maps $\hat{o} \in \{0, 1\}^{|T| \times H \times W}$ as Eq. \ref{eqn:binary_prediction}.
\begin{equation}
    \label{eqn:binary_prediction}
    \hat{o} = Threshold({\hat{y}_N}, \theta),
\end{equation}
where $\theta = 0.5$ in this study. Then, we computed the Dice score of a region $t \in T$ as Eq. \ref{eqn:dice_score}.

\begin{equation}
    \label{eqn:dice_score}
    DiceScore_t = { 2*\sum_{j=1}^{H \times W} {\hat{o}[t][j] \cdot y[t][j]} \over \sum_{j=1}^{H \times W}{\hat{o}[t][j]} + \sum_{j=1}^{H \times W}{y[t][j]} },
\end{equation}
where $y$ is the ground truth segmentation map. We decided the correctness of a pixel $j$ according to the following criteria:
\begin{equation}
   \begin{cases}
        \textbf{true positive} \ \text{if} \ \hat{o}[t][j] = 1 \text{ and } y[t][j] = 1 \\
        \textbf{false positive} \ \text{if} \ \hat{o}[t][j] = 1 \text{ and } y[t][j] = 0 \\
        \textbf{true negative} \ \text{if} \ \hat{o}[t][j] = 0 \text{ and } y[t][j] = 0 \\        
        \textbf{false negative} \ \text{if} \ \hat{o}[t][j] = 0 \text{ and } y[t][j] = 1 \\
   \end{cases}
\end{equation}

\subsection{Bolus Segmentation Performance}

We evaluated the bolus segmentation performance of PECI-Net. For comparison, we also evaluated five baseline models using the same data. The baseline models include UNet \citep{unet}, TernausNet \citep{ternausnet}, TransUNet \citep{TransUNet}, HiFormer \citep{hiformer}, and SwinUNet \citep{swin-unet}. For fair evaluation, we first set the hyperparameters of the baseline models to the default values provided by their authors and then further tuned them to improve performance in bolus segmentation.

\begin{table*}[h]
{
\centering
\caption{Comparison with Previous Studies}
\label{tab:segmentation_performance}
\resizebox{\textwidth}{!}{%
\begin{tabular}{llllllll}
\hline
Model           & Bolus           & Cervical Spine  & Hyoid Bone      & Mandible        & Soft Tissue     & Vocal Fold      & Average           \\ \hline
UNet \citep{unet}            & 0.6262          & 0.9207          & 0.7726          & 0.8118          & 0.8358          & 0.7026          & 0.7783          \\
TernausNet \citep{ternausnet}      & 0.6891          & 0.9454          & 0.7930          & \textbf{0.8153} & \textbf{0.8542} & 0.6997          & 0.7995          \\
TransUNet \citep{TransUNet}      & 0.6829          & 0.9212          & 0.7954          & 0.8086          & 0.8328          & 0.7068          & 0.7913          \\
HiFormer \citep{hiformer}        & 0.6651          & 0.9431          & 0.7712          & 0.7971          & 0.8404          & 0.6779          & 0.7824          \\
SwinUNet \citep{swin-unet} & 0.6741         & \textbf{0.9543}                  & 0.7661              & 0.7905            & 0.8499               & \textbf{0.7185}              & 0.7922  \\
PECI-Net (Ours) & \textbf{0.7345} & 0.9459 & \textbf{0.8051} & 0.8060          & 0.8538          & 0.7183 & \textbf{0.8106} \\ \hline
\end{tabular}%
}
}
\end{table*}

Table \ref{tab:segmentation_performance} presents our main results.
For bolus, the primary target of this study, PECI-Net showed an accuracy of 73.45\%, which is 10.83\% higher than UNet and 4.54\% higher than the second best model, TernausUNet.
Despite the fact that PECI-Net was trained with a focus on bolus segmentation as described in Subsection \ref{subsec:training}, PECI-Net showed the highest average accuracy of 81.06\%, which is 1.11\% higher than the second best model, TernausNet. Besides the bolus, PECI-Net showed the highest Dice scores for hyoid bone. TernausNet performed the best for the mandible and soft tissue and SwinUNet for the cervical spine and vocal fold.

\subsection{Ablation Studies}

We conducted ablation studies to analyze the effect of the proposed methods. We built two baseline models for comparative evaluation. The first baseline model consists of CIN but does not apply PEN. This model takes the original VFSS image as input and segments the bolus and other target regions using CIN. Since the VFSS image has only a single channel, we replicated its gray-scale values to convert it to a three-channel image. We built the second baseline model by removing both CIN and PEN from PECI-Net. Such modification results in TransUNet as we implemented each stage of PECI-Net based on TransUNet.

Table \ref{tab:ablation_studies} presents the results of ablation studies. Removing PEN from PECI-Net decreased bolus segmentation performance by 1.89\% from 73.45\% to 71.56\%, but slightly increased average performance from 81.06\% to 81.10\%. These results suggest that the training of PEN with a focus on bolus segmentation improves bolus segmentation performance but can slightly sacrifice performance for other regions.
When we replaced CIN with vanilla TransUNet, bolus segmentation performance further decreased by 3.27\%. In this case, the average performance was also decreased by 1.97\%, suggesting that referring to context helps the segmentation of all regions. Consequently, the results of the ablation studies confirm that both PEN and CIN are effective in improving bolus segmentation performance.

\begin{table*}[h]
{
\centering
\caption{The Results of Ablation Studies}
\label{tab:ablation_studies}
\resizebox{\textwidth}{!}{%
\begin{tabular}{llllllll}
\hline
Model  & Bolus & Cervical Spine & Hyoid Bone & Mandible & Soft Tissue & Vocal Fold & Average \\ \hline
TransUNet \citep{TransUNet}      & 0.6829         & 0.9212                  & 0.7954              & 0.8086            & 0.8328               & 0.7068              & 0.7913           \\
CIN (w/o PEN)   & 0.7156         & \textbf{0.9501}                  & 0.7863              & \textbf{0.8123}            & \textbf{0.8671}               & \textbf{0.7345}              & \textbf{0.8110}           \\
PECI-Net (Ours) & \textbf{0.7345}         & 0.9459                  & \textbf{0.8051}              & 0.8060            & 0.8538               & 0.7183              & 0.8106          \\ \hline
\end{tabular}%
}
}
\end{table*}

\subsection{Performance by the Number of Segmentation Stages}
\label{subsec:no_stages}
To analyze the effect of the number of stages on performance, we varied the number of stages and measured bolus segmentation performance. The results are presented in Table \ref{tab:performance_by_stages}. The 2-stage model showed the highest performance of 73.45\%, which is 2.41\% higher than the single-stage model. The 3-stage model showed the second highest performance of 73.23\%, which is slightly lower than the 2-stage model. The 4-stage model showed a poor performance. However, on the training data, the performance of the 4-stage model showed an accuracy of 89.50\%, which is 0.7\% higher than that of the 2-stage model, 88.80\%, suggesting that employing too many stages for a limited dataset can lead to overfitting. The inference time increases linearly with the number of stages. The average time per stage is about 8 msec.

\begin{table*}[h]
  {
    \centering
    \caption{Performance by the number of segmentation stages}
    \label{tab:performance_by_stages}
    \resizebox{\textwidth}{!}{%
      \begin{tabular}{llllllll}
        \hline
        Model & Bolus & Cervical Spine & Hyoid Bone & Mandible & Soft Tissue & Vocal Fold & Average \\ \hline
        PECI-Net (1 stage) & 0.7104 & 0.9468 & 0.7854 & 0.8013 & 0.8450 & 0.6949 & 0.7973 \\
        PECI-Net (2 stages) & \textbf{0.7345} & 0.9459 & \textbf{0.8051} & 0.8060 & 0.8538 & 0.7183 & \textbf{0.8106} \\
        PECI-Net (3 stages) & 0.7323 & 0.9528 & 0.7827 & \textbf{0.8228} & 0.8464 & \textbf{0.7193} & 0.8094 \\
        PECI-Net (4 stages) & 0.6976 & \textbf{0.9553} & 0.7880 & 0.8055 & \textbf{0.8543} & 0.7155 & 0.8027 \\ \hline
      \end{tabular}%
    }
  }
\end{table*}

\subsection{Performance According to Architecture}
To evaluate the performance according to the architecture of the segmentation stages, we created variants of PECI-Net applying UNet \citep{unet} and SwinUNet \citep{swin}, in addition to TransUNet \citep{TransUNet}, to each stage and compared them with their base models. The three base models have similar symmetrical structures but differ in the building blocks. UNet comprises convolution blocks, while SwinUNet consists of SwinTransformer blocks. TransUNet has a hybrid architecture that combines convolution and transformer blocks.
The results are presented in Table \ref{tab:performance_arch}. All three PECI-Net variants showed higher performance than the corresponding base models. The PECI-Net based on UNet showed an accuracy of 63.32\%, which is 0.7\% higher than its base model. SwinUNet's performance was 67.41\%, but when extended to PECI-Net, it performed 68.53\%, showing an improvement of 1.12\%. TransUNet showed the most significant improvement of 5.16\% when extended to PECI-Net, from 68.29\% to 73.45\%.

\begin{table*}[h]
{
\centering
\caption{Performance according to architecture}
\label{tab:performance_arch}
\resizebox{\textwidth}{!}{%
\begin{tabular}{llllllll}
\hline
Model  & Bolus & Cervical Spine & Hyoid Bone & Mandible & Soft Tissue & Vocal Fold & Average \\ 
\hline

TransUNet \citep{TransUNet}     & 0.6829          & 0.9212          & 0.7954          & 0.8086          & 0.8328          & 0.7068          & 0.7913           \\
PECI-Net based on TransUNet & \textbf{0.7345} & 0.9459 & \textbf{0.8051} & 0.8060          & \textbf{0.8538}          & 0.7183 & \textbf{0.8106}           \\
\hline

UNet \citep{unet}      & 0.6262         & 0.9207                  & 0.7726              & \textbf{0.8118}            & 0.8358               & 0.7026              & 0.7783           \\
PECI-Net based on UNet   & 0.6332         & 0.9340                  & 0.7316              & 0.7723            & 0.8280               & 0.6763              & 0.7626           \\
\hline

SwinUNet \citep{swin-unet} & 0.6741         & \textbf{0.9543}                  & 0.7661              & 0.7905            & 0.8499             & 0.7185              & 0.7922          \\
PECI-Net based on SwinUNet   & 0.6853          & 0.9536                  & 0.7752              & 0.7971            & 0.8436               & \textbf{0.7377}              & 0.7987           \\ \hline
\end{tabular}%
}
}
\end{table*}

\subsection{Comparison of Preprocessing Algorithms}

We evaluated the effect of the preprocessing ensemble by comparing PECI-Net with baseline models that combine CIN with the five preprocessing algorithms listed in Subsection \ref{subsec:PEN}. Table \ref{tab:preprocessing_comparison} shows the results. Despite the fact that each of the five processing algorithms enhances the VFSS image in a specific aspect, none of them brought a meaningful improvement in bolus segmentation when applied alone. However, when they were combined together by PEN, they brought a significant improvement of 1.89\% in bolus segmentation. These results support that PEN successfully learns the optimal combination of preprocessing algorithms for the target task and therefore is effective in improving performance in bolus segmentation.

\begin{table*}[h]
{
\centering
\caption{Bolus segmentation performance by preprocessing algorithm}
\label{tab:preprocessing_comparison}
\resizebox{\textwidth}{!}{%
\begin{tabular}{llllllll}
\hline
Preprocessing & Bolus & Cervical Spine & Hyoid Bone & Mandible & Soft Tissue & Vocal Fold & Average \\ \hline
Identity mapping       & 0.7156         & 0.9501         & 0.7863              & \textbf{0.8123}            & \textbf{0.8671}               & \textbf{0.7345}              & \textbf{0.8110}         \\
Laplacian sharpening   & 0.7070         & 0.9454         & 0.7986              & 0.7819            & 0.8428               & 0.7090              & 0.7974         \\
CLAHE                  & 0.6965         & 0.9476         & 0.7918              & 0.8023            & 0.8597               & 0.7036              & 0.8002         \\
Double CLAHE           & 0.7135         & \textbf{0.9553}         & 0.7644              & 0.7932            & 0.8522               & 0.6968              & 0.7959         \\
CLAHE + Sharpening     & 0.7166         & 0.9505         & 0.7981              & 0.8022            & 0.8528               & 0.7316              & 0.8086         \\
PEN (Ours)             & \textbf{0.7345}         & 0.9459         & \textbf{0.8051}              & 0.8060            & 0.8538               & 0.7183              & 0.8106         \\ \hline
\end{tabular}%
}
}
\end{table*}

We also measured the speed of the preprocessing algorithms. The identity mapping does not require additional time. The average execution time of Laplacian sharing and CLAHE was 2.11 msec and 0.51 msec, while Double CLAHE and `CLAHE + Sharpening' took 0.91 msec and 4.07 msec, respectively. The total time required to run all four preprocessing algorithms was only 7.60 msec per image.

\subsection{Performance by the Number of Preprocessing Algorithms}
We also measured the performance of PEN by the number of preprocessing algorithms. We incrementally added preprocessing algorithms and measured the performance. The first model has only identity mapping, effectively applying no preprocessing. We then sequentially added Laplacian sharpening and CLAHE to the second and third models. Finally, we added compound preprocessing algorithms, `CLAHE + Sharpening' and Double CLAHE, in the fourth and fifth models.

Table \ref{tab:performance_no_preprocessing} presents the results. Bolus segmentation accuracy increased with the number of preprocessing algorithms. Applying only the Laplacian sharpening resulted in a slight improvement of 0.04\%. Adding CLAHE, `CLAHE + Sharpening', and Double CLAHE led to improvements of 0.33\%, 0.08\%, and 1.44\%, respectively. We also ran an additional experiment that applies the three CLAHE-based algorithms but not identity mapping. This model showed a performance of 72.00\%, which is 1.45\% lower than our best model. These results suggest that all four preprocessing algorithms effectively supplement the input image, but they miss some information in the input image.

\begin{table*}[h]
{
\centering
\caption {Performance according to the number of preprocessing algorithms}
\label{tab:performance_no_preprocessing}
\resizebox{\textwidth}{!}{%
\begin{tabular}{llllllll}
\hline
Preprocessing & Bolus & Cervical Spine & Hyoid Bone & Mandible & Soft Tissue & Vocal Fold & Average \\ \hline
Identity mapping (CIN)       & 0.7156         & 0.9501                  & 0.7863              & 0.8123            & \textbf{0.8671}               & \textbf{0.7345}              & 0.8110         \\
Identity mapping \& Laplacian sharpening    & 0.7160         & 0.9515                  & 0.7943              & 0.8029            & 0.8499               & 0.7186              & 0.8055         \\
Identity mapping \& Laplacian sharpening \& CLAHE                  & 0.7193         & \textbf{0.9551}                  & \textbf{0.8052}              & \textbf{0.8142}            & 0.8611               & 0.7330             & \textbf{0.8146}         \\
Identity mapping \& Laplacian sharpening \& CLAHE \& CLAHE + Sharpening            & 0.7201         & 0.9470                  & 0.7978              & 0.8008            & 0.8556               & 0.7222              & 0.8072         \\
Identity mapping \& Laplacian sharpening \& CLAHE \& CLAHE + Sharpening \& Double CLAHE             & \textbf{0.7345}         & 0.9459                  & 0.8051              & 0.8060            & 0.8538               & 0.7183              & 0.8106         \\
 \hline
\end{tabular}%
}
}
\end{table*}
 
\subsection{Performance According to the Choice of Regions Used as Context}

We evaluated the performance according to the choice of regions used as the context for the second segmentation stage. We constructed four PECI-Net variants, and the second stage of each variant refers to a different combination of regions among the first-stage segmentation results.
The second stage refers to the first-stage segmentation maps of all regions in the first model but refers only to those of the cervical spine and mandible in the second model. The context for the second stage in the third and fourth models consists of the regions not selected in the proposed model: the hyoid bone and vocal fold in the third model and soft tissue in addition to these two regions in the fourth model.

Table \ref{tab:performance_choiceofregions} shows the results. The first model showed an accuracy of 72.05\%. The second model showed the best performance of 73.45\%, which is 1.4\% higher than the first model. However, the third and fourth models, which include only dimmed or small regions in the context for the second stage, showed lower performances of 70.38\% and 71.25\%, respectively. Especially, the performance of the third model was even lower than that of the single-stage model, 71.04\% in Table \ref{tab:performance_by_stages}. These results suggest that the choice of context affects performance significantly.

\begin{table*}[h]
{
\centering
\caption {The performance of PECI-Net according to the choice of regions used as context}
\label{tab:performance_choiceofregions} 
\resizebox{\textwidth}{!}{%
\begin{tabular}{llllllll}
\hline
Selected Regions & Bolus & Cervical Spine & Hyoid Bone & Mandible & Soft Tissue & Vocal Fold & Average \\ \hline
All regions                  & 0.7205         & 0.9501                  & 0.7922              & 0.8023            & \textbf{0.8588}               & \textbf{0.7240}             & 0.8080         \\

Cervical Spine and Mandible (Ours)             & \textbf{0.7345}         & 0.9459                  & \textbf{0.8051}              & 0.8060            & 0.8538               & 0.7183              & \textbf{0.8106}         \\

Hyoid bone and Vocal fold      & 0.7038         & \textbf{0.9550}                  & 0.7908              & 0.7881            & 0.8512               & 0.7112              & 0.8000         \\

Hyoid bone, Vocal fold, and Soft tissue & 0.7125         & 0.9525                  & 0.7760              & \textbf{0.8113}            & 0.8533               & 0.7247              & 0.8051         \\

\hline
\end{tabular}%
}
}
\end{table*}

\subsection{Qualitative Evaluation and Discussion}

\begin{figure*}[p]
    \centering
    \includegraphics[width=0.8\textwidth]{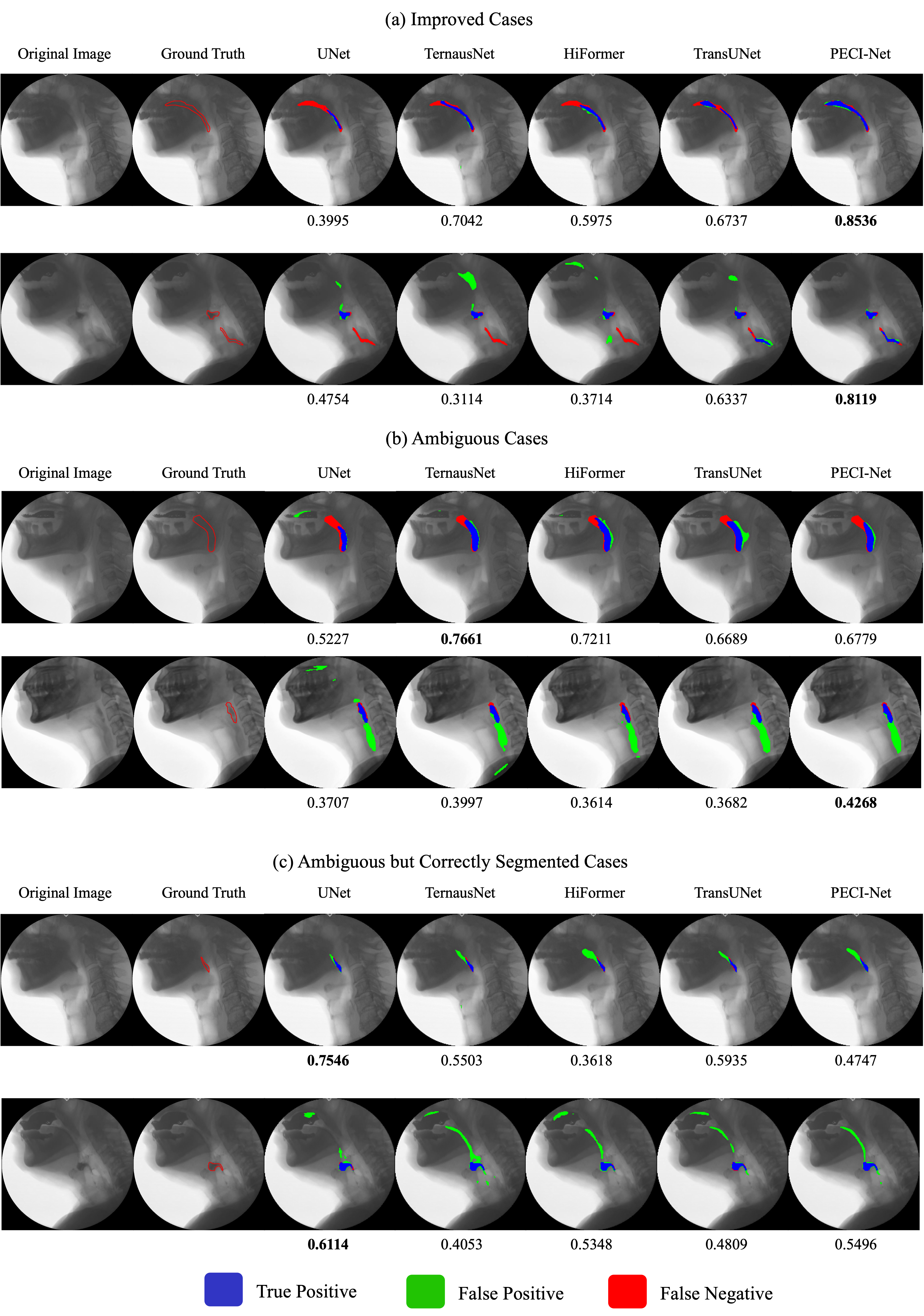}
    \caption{Bolus segmentation results of PECI-Net and four baseline models. The numbers represent the Dice scores.
    (a) VFSS images that PECI-Net segmented more accurately than baseline models
    (b) ambiguous VFSS images that are hard to correctly segment even with human eyes
    (c) ambiguous VFSS images for which PECI-Net produced results that differed from the ground truth bolus regions but were agreed upon by another rater.
     }
    \label{fig:predictions_comparison}
\end{figure*}

Fig. \ref{fig:predictions_comparison} displays bolus segmentation results of PECI-Net and four baseline models. Fig. \ref{fig:predictions_comparison}(a) shows the VFSS images that PECI-Net segmented more accurately than the baseline models. The bolus regions in these images are vague and unclear, and therefore the baseline models produced many false positive (green) and false negative (red) estimations. However, PECI-Net exhibited remarkably improved results.

Fig. \ref{fig:predictions_comparison}(b) shows ambiguous samples in which the boundaries of the bolus regions are unclear. Both PECI-Net and the baseline models produced significant areas of false negatives in the top row and false positives in the bottom row. However, such samples are hard to correctly segment even with human eyes and can be annotated differently depending on the rater.
Fig. \ref{fig:predictions_comparison}(c) presents ambiguous but correctly segmented samples. During data collection, the original rater conservatively annotated these samples to narrowly determine the bolus region. However, when another rater checked the segmentation results of the model, he determined that the green areas were also bolus regions.

\begin{figure*}[tb]
    \centering
    \includegraphics[width=0.7\textwidth]{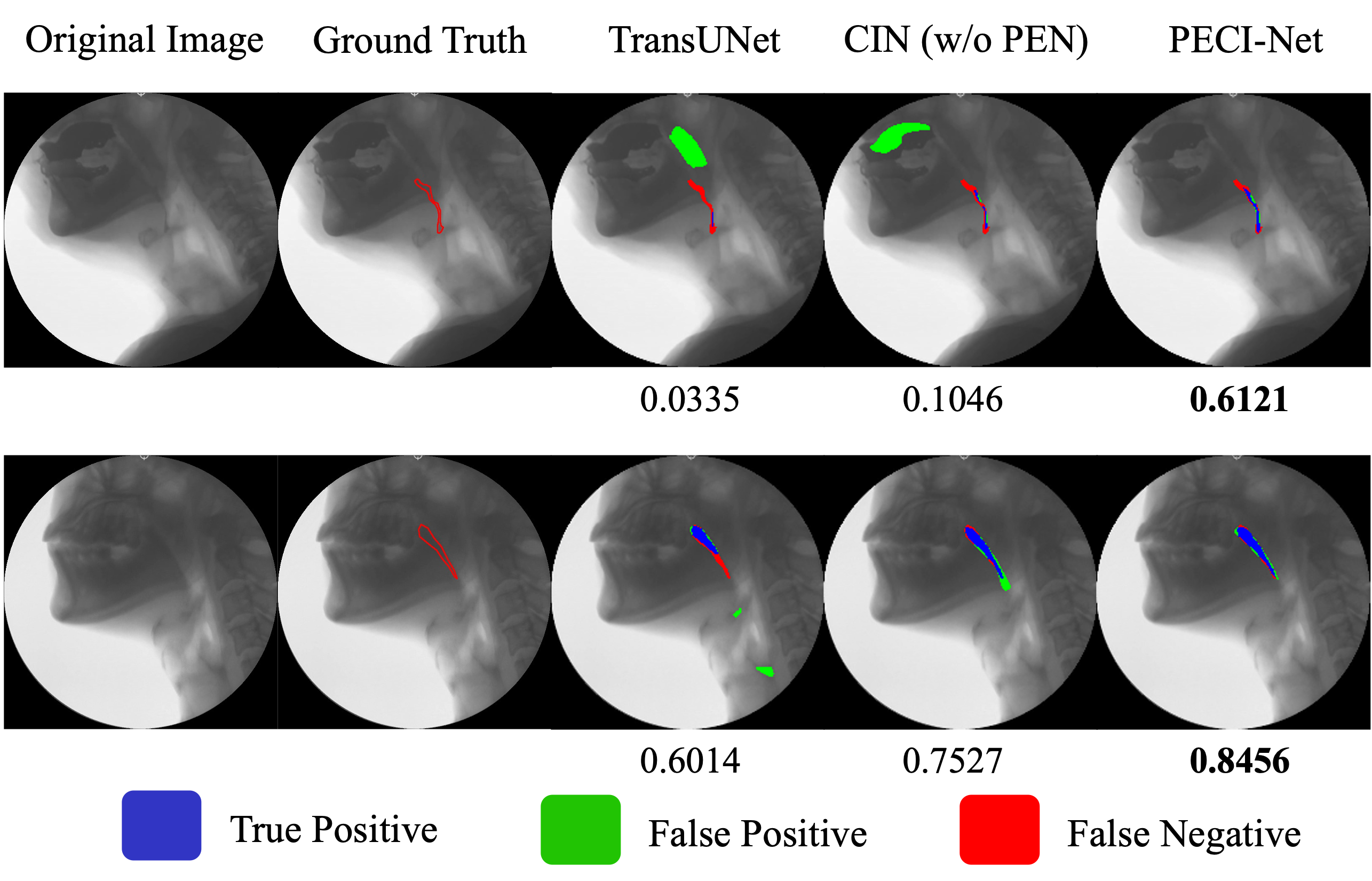}
    \caption{The segmentation results of the models used in the ablation studies (TransUNet, CIN, and PECI-Net).
    }
    \label{fig:peci_comparison}
\end{figure*}

When the bolus area overlapped with dark areas such as the mandible and hard palate, the boundaries of the bolus area were fuzzy, increasing the segmentation error. Another major source of error was the presence of dentures. Dentures appear dark in VFSS images, and PECI-Net sometimes segmented the area of the dentures as the bolus.
We suspect that this is because our dataset does not contain the annotation of denture regions, and therefore PECI-Net could not learn to separate bolus from dentures.
To overcome this limitation, it is necessary to collect VFSS images containing regions that are easily confused with bolus and to label those regions.

Fig. \ref{fig:peci_comparison} shows the bolus regions estimated by the three models compared in the ablation studies. TransUNet produced poor segmentation results for these images. However, CIN segmented the bolus region more accurately by using contexts from the cervical spine and mandible regions. PECI-Net further improved segmentation accuracy by incorporating PEN and CIN.

\section{Conclusion}

Bolus segmentation is crucial for accurate and automatic diagnosis of swallowing disorders. However, segmenting bolus regions in a VFSS image accurately is challenging because of the non-fixed shape of the bolus, the translucency and insufficient quality of VFSS images, and the absence of color information. To overcome these challenges, we present PECI-Net, a novel architecture specialized for VFSS image segmentation. PECI-Net consists of a preprocessing ensemble network (PEN), which combines multiple preprocessing algorithms in a learnable manner, and a cascaded inference network (CIN), which segments bolus and other regions accurately by referencing the spatial relation between the bolus and organs.
To train and evaluate the proposed methods, we collected 2155 VFSS images and manually labeled the region of bolus, cervical spine, hyoid bone, mandible, soft tissue, and vocal fold.
In experiments, PECI-Net showed a bolus segmentation accuracy of 73.45\%, which is 10.83\% higher than the widely used UNet and 4.54\% higher than the second best model, TernausNet, demonstrating the effectiveness of the proposed methods.
Future studies with the results of this study are needed to develop an automated and objective VFSS reading system in the clinical field.

\section*{Acknowledgements}
This research was supported by Pohang Stroke and Spine Hospital and the MSIT (Ministry of Science and ICT), Korea, under the National Program for Excellence in SW) supervised by the IITP (Institute of Information \& Communications Technology Planning \& Evaluation) in 2023 (2023-0-00055).

\bibliographystyle{plainnat}
\bibliography{reference}

\end{document}